\def\year{2020}\relax
\documentclass[letterpaper]{article} % DO NOT CHANGE THIS
\usepackage{aaai20}  % DO NOT CHANGE THIS
\usepackage{times}  % DO NOT CHANGE THIS
\usepackage{helvet} % DO NOT CHANGE THIS
\usepackage{courier}  % DO NOT CHANGE THIS
\usepackage[hyphens]{url}  % DO NOT CHANGE THIS
\usepackage{graphicx} % DO NOT CHANGE THIS
\usepackage{graphicx}  % DO NOT CHANGE THIS
\usepackage{multirow}
\usepackage{amsmath}
\usepackage{amssymb}

\title{Object-Guided Instance Segmentation for Biological Images}
\author{Jingru Yi\textsuperscript{\rm 1}\thanks{This work was funded partly by NSF 1763523, 1747778, 1733843, 1703883 grants.}, Hui Tang\textsuperscript{\rm 2},  Pengxiang Wu\textsuperscript{\rm 1}, Bo Liu\textsuperscript{\rm 1}, Daniel J. Hoeppner\textsuperscript{\rm 3},\\ \Large \textbf{Dimitris N. Metaxas\textsuperscript{\rm 1}, Lianyi Han\textsuperscript{\rm 2}, Wei Fan\textsuperscript{\rm 2}}\\  % All authors must be in the same font size and format. Use \Large and \textbf to achieve this result when breaking a line
\textsuperscript{\rm 1}Department of Computer Science, Rutgers University, Piscataway, NJ 08854, USA\\
\textsuperscript{\rm 2}Tencent Hippocrates Research Labs, Palo Alto, CA94306, USA\\
\textsuperscript{\rm 3}Lieber Institute for Brain Development, MD 21205, USA\\
jy486@cs.rutgers.edu
}
 
\begin{document}

\maketitle

\begin{abstract}
Instance segmentation of biological images is essential for studying object behaviors and properties. The challenges, such as clustering, occlusion, and adhesion problems of the objects, make instance segmentation a non-trivial task. Current box-free instance segmentation methods typically rely on local pixel-level information. Due to a lack of global object view, these methods are prone to over- or under-segmentation. On the contrary, the box-based instance segmentation methods incorporate object detection into the segmentation, performing better in identifying the individual instances. In this paper, we propose a  new box-based instance segmentation method. Mainly, we locate the object bounding boxes from their center points. The object features are subsequently reused in the segmentation branch as a guide to separate the clustered instances within an RoI patch. Along with the instance normalization, the model is able to recover the target object distribution and suppress the distribution of neighboring attached objects. Consequently, the proposed model performs excellently in segmenting the clustered objects while retaining the target object details. The proposed method achieves state-of-the-art performances on three biological datasets: cell nuclei, plant phenotyping dataset, and neural cells.
\end{abstract}

\section{Introduction}
Instance segmentation is a task that assigns the instance labels to every pixel of the input images. In biological images, instance segmentation is a fundamental step in analyzing the object behaviors and properties, such as cell interaction, nuclei treatment reaction, and plant phenotyping. Instance segmentation of biological images is challenging due to the object clustering, adhesion, and occlusion. Besides, the biological tasks usually require capturing the fine details of the instances, such as the leaf stalking and cell protrusions.  A fast and accurate instance segmentation tool will benefit a lot to the biological society.

Existing methods of instance segmentation can be divided into two types: box-free instance segmentation and box-based instance segmentation. Box-free instance segmentation methods segment the object instances by analyzing the instance morphology properties (e.g., object contours, textures, and shapes) without the aid of object bounding boxes. Box-free instance segmentation methods usually suffer from separating the touching objects. For example, DCAN \cite{chen2016dcan} and Deep Watershed \cite{bai2017deep} are prone to over-segmentation in the scenarios of unclear object boundaries. Cosine Embedding \cite{payer2018cosine} separates the touching objects through pixel embedding clustering, but it tends to generate fragmentary segmentation. Some other methods, such as StarDist \cite{schmidt2018cell}, try to solve the clustering problem using object shape information, yet their application is limited to convex-shape objects.

% Figure 1  %%%%%%%%%%%%%%%%%%%%%%%%%%%%%%%%%%%%%%%%%
\begin{figure*}[tbh!]
\centering
\includegraphics[width=0.9\textwidth]{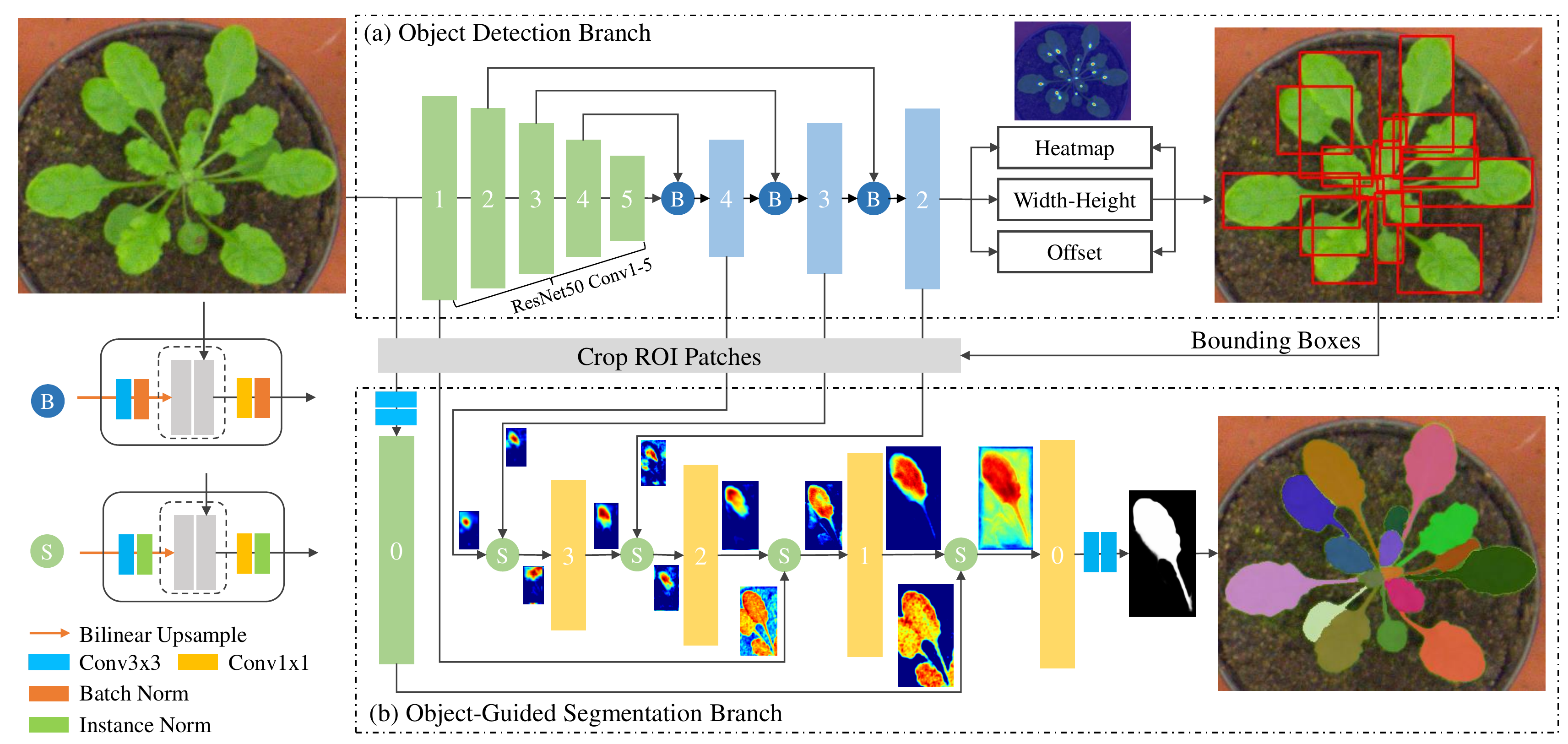}
\caption{Object-guided instance segmentation framework. The encoder (layers1-5) is from a ResNet50 \cite{he2016deep} network. The framework contains two branches: (a) object detection branch and (b) object-guided segmentation branch. The B and S represent the skip combination modules. The object detection branch predicts the center heatmaps, box properties (e.g., width and height), and center offsets. The predicted bounding boxes are flowed to the segmentation branch to crop RoI patches. The object features are exploited as guidance for the model to separate the attached objects.}
\label{fig1}
\end{figure*}
%%%%%%%%%%%%%%%%%%%%%%%%%%%%%%%%%%%%%%%%%%%%%%%%%%%%%%

Box-based instance segmentation methods combine object detection and segmentation. They locate the bounding boxes of the objects from a global perspective and subsequently refine the instance segmentation within a cropped region of interest (RoI) patch.  An accurate object detector plays a crucial role in the box-based instance segmentation methods. The current box-based instance segmentation methods \cite{he2017mask,yi2019attentive,yi2019context} generally adopt the anchor box-based object detectors, which spread the anchor boxes densely on the feature maps and predict the offsets to the anchor boxes. However, the anchor box-based object detector suffers from a severe imbalance issue between the positive and negative anchor boxes, which would result in slow training and sub-optimal detection performances \cite{law2018cornernet}. Also, the RoIAlign-based methods, such as Mask R-CNN \cite{he2017mask}, predict coarse segmentation masks for objects within the fixed-size RoI patches (e.g., $14\times14$), losing the fine details of the objects. Keypoint Graph \cite{YI2019MICCAI} proposes to use the keypoint-based object detector and a separate segmentation branch to solve these issues. When achieving better performance, it fails to localize the small objects where the five keypoint circles of a bounding box would overlap. To deal with this problem, several works \cite{ribera2019locating,zhou2019objects} suggest locating objects through their center points. The center keypoint detection allows more small objects to be localized. Besides, detecting only one keypoint for each instance removes the complex grouping process as in the Keypoint Graph, which makes the model computationally efficient.

In this work, we propose a new keypoint-based instance segmentation method. In particular, we localize the objects through their center points. The bounding box properties (e.g., width and height) are obtained from the center points. The center object features are subsequently reused with instance normalization to help the model focus on the target objects. In this manner, our model is able to separate the target from its neighboring attached objects and preserve its fine details. Our contributions are summarized as follows:
\begin{itemize}
    \item We design a novel object-guided architecture for accurate and fast instance segmentation of biological images. 
    \item The proposed model reuses the center object features in the segmentation branch to separate the attached objects.
    \item Along with the instance normalization, the proposed model is able to suppress the distribution of neighbor objects and focus on the target.
    \item Experimental results demonstrate the superior performance of the proposed model to the state-of-the-arts in terms of accuracy and efficiency.
\end{itemize}

\section{Related Work}
In this section, we briefly review two main categories of instance segmentation methods: box-free and box-based instance segmentations.

\subsection{Box-free Instance Segmentation}
Box-free instance segmentation methods segment the objects by analyzing the local pixel-level features of the images. For example, DCAN \cite{chen2016dcan} fuses the predicted contour map with the semantic segmentation map to separate the connected objects at the cost of losing the boundary pixels. Deep Watershed \cite{bai2017deep} learns the instance energy map and cuts the instances according to an energy threshold. DCAN and Deep Watershed are subject to the quality of object boundary features. Cosine Embedding \cite{payer2018cosine} embeds the image pixels to a high dimensional space. It then obtains the instance masks by clustering the pixel embeddings. Due to the failure of clustering, Cosine Embedding usually generates incomplete instance fragments. StarDist \cite{schmidt2018cell} utilizes the star-convex polygons to describe the object shapes. However, it applies only to objects with convex shapes. 

\subsection{Box-based Instance Segmentation}
Box-based instance segmentation comprises both object detection and object segmentation. It first localizes the objects using bounding boxes and then segments the objects within the cropped RoI patches. One representative work is Mask R-CNN \cite{he2017mask}. It incorporates a mask branch into the FPN \cite{lin2017feature} network for instance segmentation. In addition, it proposes a RoIAlign method to extract RoI patches with fixed size for simultaneous location regression, classification, and mask prediction. However, the fixed-size of RoI patches can hardly capture the instance details, such as the protrusion of neural cells. Besides, the imbalanced positive and negative anchor boxes would incur slow training and sub-optimal detection performance. A good object detector plays a key role in the box-based instance segmentation. Most recently, keypoint-based object detectors, such as CornerNet \cite{law2018cornernet} and ExtremeNet \cite{zhou2019bottom}, are developed to solve the imbalance problem of anchor-box based detectors. In the spirit of detecting keypoints, Keypoint Graph \cite{YI2019MICCAI} proposes to detect the four corners and the center point of a bounding box. It identifies an instance by grouping the five keypoints according to a keypoint graph. One weakness of this method is that, since each keypoint circle has a fixed radius, the five keypoints tend to overlap for small objects. Consequently, the Keypoint Graph suffers from the failure of detecting small objects. In this work, we only detect the center points of the objects, which are beneficial to the localization of much smaller objects.

\section{The Proposed Method}
The framework of our method is illustrated in Fig.~\ref{fig1}. The object-guided instance segmentation comprises two branches: object detection and object-guided segmentation. The object detection branch aims to provide the bounding boxes of instances. The detected bounding boxes are then employed to crop the RoI patches from the input feature maps of the segmentation branch. Instance segmentation is subsequently performed on these RoI patches.

\subsection{Object Detection Branch}
Detecting objects with a grouping of keypoints would fail for small objects since the box keypoints would overlap in such scenarios. On the contrary, a single center point does not require the keypoint grouping process and therefore is more suitable for identifying small objects. In this paper, we localize the objects directly through their center points. We use conv1-5 layers from a ResNet50 \cite{he2016deep} as an encoder to extract features. The detection branch combines the deep features with the shallow ones through a skip connection \cite{ronneberger2015u}. The output of the object detection branch comprises three parts: a center heatmap, a center offset map, and a width-height map. Note that the object detection branch is not a mirrored architecture because the model performed worse on mirrored architecture from our experiments ($\sim$2-3 points lower). The reason would be that on a keypoint heatmap, the imbalance issue between the foreground and background pixels gets more severe on a full-size feature map as the number of positive objects is constant. Also, a downsized output can speed up the network because the number of parameters is smaller compared to a full-size output.

\subsubsection{Center Heatmap.}
The center heatmap is a key module of the object detection branch. For each object, there is only one ground-truth positive location on the input image. Following the work of \cite{law2018cornernet}, instead of penalizing all the background pixels, we reduce the penalty around a Gaussian circle of each ground-truth center point. We use the variant focal loss to optimize the parameters:
\begin{equation}
    L_{hm} =
    -\frac{1}{N}
    \begin{cases}(1-p_i)^\alpha \log(p_i)&\text{if }  y_i=1\\
    (1-y_i)^\beta(p_i)^\alpha\log(1-p_i)&\text{otherwise}
    \end{cases},
\end{equation}
where $i$ indexes the $i$th location in the predicted heatmap, $N$ is the total number of center points, $y$ is the ground-truth. We use $\alpha=2, \beta=4$ \cite{lin2017focal,law2018cornernet} in this paper. The predicted center heatmaps are refined through a non-maximum-suppression (NMS) operation. The operation employs a $3\times 3$ max-pooling layer on the center heatmaps. The center points are gathered according to their local maximum probability.

\subsubsection{Offset Map.}
As the center location does not rely on the detailed morphology information, to reduce the computational cost, the keypoints are usually predicted in a downsized heatmap \cite{law2018cornernet,newell2016stacked}. An offset map is necessary to map the center locations back to the original image correctly. Suppose $n$ is the downsized factor, $(x,y)$ is a location in the input image. The offset map can be represented as:
\begin{equation}
    o_i = (\frac{x_i}{n}-\lfloor\frac{x_i}{n}\rceil, \frac{y_i}{n}-\lfloor\frac{y_i}{n}\rceil),
\end{equation}
where $i$ indexes the $i$th center point. In the training process, we apply the L1 loss to regress the offset at the center points as it is more resistant to outliers.

\subsubsection{Width-Height Map.}  Different from anchor-box based detectors, the proposed model directly regresses the width and height of the bounding boxes from the center points. Similar to the offset map, we use the L1 loss to regress the width and height of the bounding boxes at the center points.

% Figure 2 %%%%%%%%%%%%%%%%%%%%%%%%%%%%%%%%
\begin{figure}[t]
\centering
\includegraphics[width=1\columnwidth]{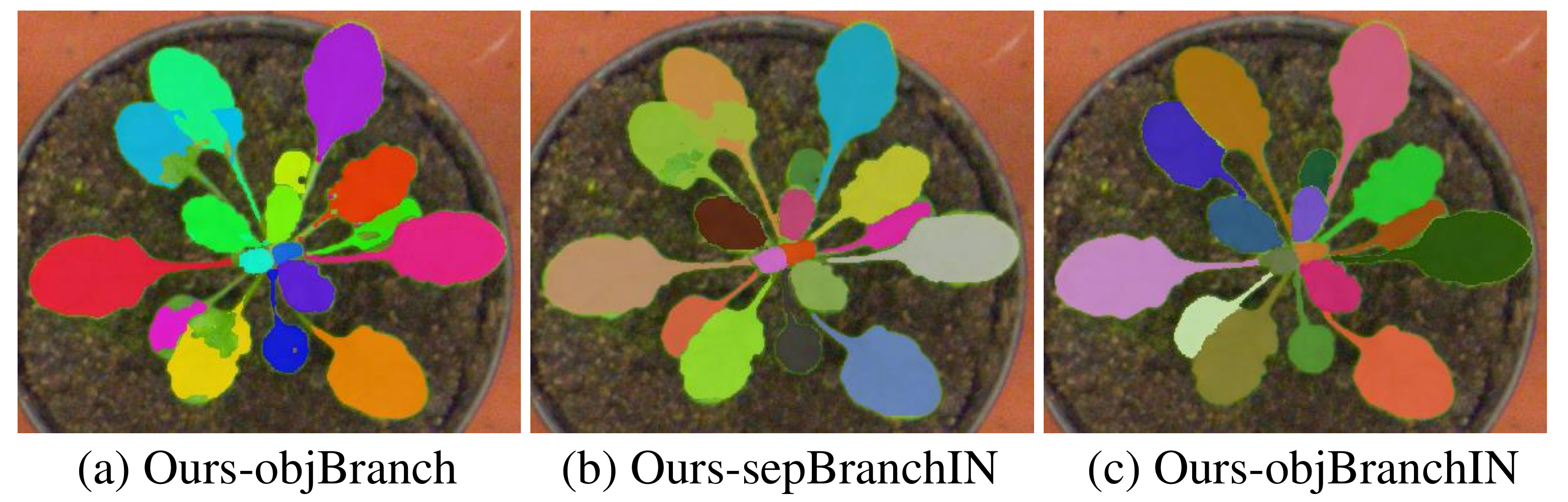}
\caption{Ablation studies of the segmentation branch on plant phenotyping dataset. The symbol ``obj" indicates the segmentation branch with the object feature, ``sep" denotes the separated segmentation branch that has no object features, ``IN" refers to the instance normalization.}
\label{fig2}
\end{figure}
%%%%%%%%%%%%%%%%%%%%%%%%%%%%%%%%%%%%%%%%%%%

\subsection{Object-Guided Segmentation Branch}
Instance segmentation on cropped RoI patches is similar to the semantic segmentation. One key challenge in this task is to separate the clustered instances while keeping the fine details of the targets. One approach \cite{he2017mask} samples a small size of pixels and uses a series of convolutional layers to remove the noise information. However, this method loses the details of the object instance. Another approach \cite{YI2019MICCAI} builds a separate segmentation branch next to the detection branch. This method tends to make mistakes in segmenting the object, which shares the same textures and classes with the neighboring attached instances. As a consequence, it can hardly separate the touching instances (see Fig.~\ref{fig2}b). To solve this problem, we propose to leverage the object information from the detection branch and on this basis build an object-guided segmentation branch.

As shown in Fig.~\ref{fig1}, after obtaining the bounding boxes from the object detection branch (Fig.~\ref{fig1}a), we crop the RoI feature patches from the encoder layers 0-1 (green feature maps) and the object layers 2-4 (blue feature maps). Then we develop a deep-to-shallow segmentation branch through the skip combination module. Note that layer 5 is not considered because the objects are too small at this scale. As can be seen from Fig.~\ref{fig1}b, the shallow layers from encoder layers 0-1 contain rich morphology details such as leaf stalks. While being beneficial to recovering the target fine details, it also brings difficulty for the network to differentiate the target object. 
For this reason, we reuse the object features (layers 2-4) as guidance to help the model separate clustering objects within an RoI patch. As can be seen from Fig.~\ref{fig2}ab, the object feature effectively helps the model separated connected objects. However, it also leads to incomplete segmentation masks. To handle this problem, we need an operation that re-calibrates the features in each cropped RoI across the spatial space and removes the unnecessary neighboring distributions. Besides, this operation should also be able to collect the statistical information of the target instances and recover the morphology details.

Instance normalization \cite{ulyanov2016instance} is a perfect choice for our task. On the one hand, feature normalization operations, including batch, group, instance normalization \cite{ioffe2015batch,wu2018group}, are proved to enable stable training for large-batch or small-batch training images. Also, instance normalization is able to remove style statistics of instance for image generation \cite{ulyanov2016instance}. For our task, each cropped RoI patch is supposed to contain mainly one target. Therefore, our problem can be formalized as removing the neighbor statistics for each RoI patch. For the predicted mask, as each RoI patch has only one channel, it is naturally feasible to apply instance normalization. Given an RoI patch $x\in \mathbb{R}^{H\times W}$, the instance normalization can be written as:
\begin{equation}
    x_{h,w}^\prime = \gamma(\frac{x_{h,w}-\mu}{\sigma})+\beta,
\end{equation}
where $\mu$ and $\sigma$ are the mean and variance of the RoI patch, respectively. $\gamma$ and $\beta$ are two learned scaling factors for the network to control the extent of RoI patch normalization.

As can be seen from Fig.~\ref{fig2}c, with the combination of object feature and instance normalization, the proposed model is able to remove the statistics of neighbor objects and recover the morphology details of the target objects. We use the binary cross-entropy loss to optimize the model parameters for the segmentation task.

%%%%%%%%%%%%%%%%%%%%%%%%%%%%%%%%%%%%%%%%%%%%%%%%%%%
\begin{table*}[tbh!]
\centering
\caption{Quantitative instance segmentation results. AP$^{box}$ and AP$^{mask}$ are the averaged AP at the box and mask IoU thresholds from 0.5 to 0.95 with an interval of 0.05. In particular, the AP$^{mask}$ at mask IoU threshold of $0.5$ and $0.75$ are listed. The averaged mask IoU (AIoU$_{mask}$) at thresholds of $0.5$ and $0.75$ are also presented to compare the quality of segmentation masks. Speed (FPS: frame per second) is measured on a single NVIDIA GeForce GTX 1080 GPU. We calculate both the model inference time and post-processing time.  AP is measured using Pascal VOC2010 metric \cite{everingham2011pascal}. The symbol ``--" denotes the very slow speed ($>$1min per image).}\smallskip
\resizebox{0.98\textwidth}{!}{
\def\arraystretch{1.1}%
\begin{tabular}{l|c|c|c|c|c|c|c|c}
\hline
Method &Datasets &AP$^{box}$ &AP$^{mask}$ & AP$^{mask}_{0.5}$ & AP$_{0.75}^{mask}$ & AIoU$_{0.5}^{mask}$ & AIoU$_{0.75}^{mask}$
& FPS \\ \hline
DCAN \cite{chen2016dcan} & \multirow{5}{*}{DSB2018} &16.47&19.70&48.36&15.56&74.11&84.27 &2.67 \\
Mask R-CNN \cite{he2017mask} & &42.13&42.58&73.94&45.05&79.69&85.13&1.01\\
Cosine Embedding \cite{payer2018instance}& &2.25&3.36&14.96&0.24&63.65&80.27&--\\
Keypoint Graph  \cite{YI2019MICCAI} & &49.07&50.63&76.13&56.72&83.33&87.38&1.54\\
Ours-objBranch&&49.34&57.90&80.15&62.00&81.29&91.50&3.42\\
Ours-sepBranchIN&&45.35&60.51&85.64&64.52&86.33&91.02&3.24\\
Ours-objBranchIN&&50.41&61.14&84.85&65.14&87.07&91.47&3.22\\
%Ours&&48.23&60.58&84.14&64.87&86.98&91.36&3.33\\
\hline
DCAN \cite{chen2016dcan} & \multirow{5}{*}{Plant} &7.03&16.67&38.86&13.02&75.78&83.76 &12.99\\
Mask R-CNN \cite{he2017mask} &&47.44&46.57&81.56&49.55&78.73&84.00&5.57\\
Cosine Embedding \cite{payer2018instance}& &5.04&6.68&20.20&3.24&70.29&82.86&--\\
Keypoint Graph  \cite{YI2019MICCAI}& &50.93&49.70&82.71&51.27&81.91&87.29&1.82\\
Ours-objBranch&&54.92&70.42&90.97&76.56&88.41&92.09&5.71\\
Ours-sepBranchIN&&55.85&74.43&93.72&79.48&88.83&92.42&5.41\\
Ours-objBranchIN&&59.45&74.11&92.20&79.15&89.31&92.80&5.45\\
%Ours&&54.11&75.25&92.55&80.19&89.83&93.12&5.56\\
\hline
DCAN \cite{chen2016dcan} & \multirow{5}{*}{Neural Cell} &1.70&10.98&44.82&1.00&64.54&79.15&4.87 \\
Mask R-CNN \cite{he2017mask} & &19.73&21.43&57.65&9.84&69.59&80.56&1.13\\
Cosine Embedding \cite{payer2018instance}& &4.29&1.98&11.71&0.9&58.82&77.31&--\\
Keypoint Graph \cite{YI2019MICCAI} & &43.68&42.77&85.11&35.94&76.23&81.55&1.86\\
Ours-objBranch&&39.40&57.78&94.99&65.68&81.52&84.96&5.24\\
Ours-sepBranchIN&&42.03&56.81&94.46&64.94&81.59&85.22&5.14\\
Ours-objBranchIN&&44.34&58.26&94.33&67.24&82.08&85.59&5.11\\
%Ours&&39.84&58.79&95.46&67.82&81.95&85.52&5.40\\
\hline
\end{tabular}
}
\label{table1}
\end{table*}
%%%%%%%%%%%%%%%%%%%%%%%%%%%%%%%%%%%%%%%%%%%

\section{Experiments}
\subsection{Datasets}
We evaluate our method on three datasets that pose different challenges in instance segmentation of biological images.

\subsubsection{DSB2018.} The cell nuclei dataset DSB2018 is obtained from the training dataset of 2018 Data Science Bowl. The dataset varies in cell types, magnification, and imaging modality as it was acquired under different conditions. We randomly split the original 670 images with annotations into training (402 images), validation (134 images), and testing (134 images) datasets.

\subsubsection{Plant Phenotyping.} The plant phenotyping dataset \cite{MinerviniPRL2015,PlantPhenotypingDatasets2015} contains 473 top-down view plant images with various image sizes. We use 284 images for training, 95 images for validation, and 94 images for testing.

\subsubsection{Neural Cell.} The neural cell dataset is sampled from a collection of time-lapse microscopic videos of rat CNS stem cells. It contains 644 gray-scale images with image size of 512$\times$640. We randomly select 386 images for training, 129 images for validation, and 129 images for testing. The neural cells have extremely irregular shapes.

\subsection{Implementation Details}
The training images are augmented using random cropping and random horizontal/vertical flipping. We set 100 epochs for training. We stop the network when the validation loss does not significantly decrease. The input resolution of training and testing images is $512\times 512$. The weights of the backbone network are pre-trained on ImageNet dataset. Other weights of the network are initialized from a standard Gaussian distribution.  We use Adam with an initial learning rate of 1.25e-4 to optimize the model weights. We implement the model with PyTorch on NVIDIA M40 GPUs.

\subsection{Evaluation Metrics}
We use the Average Precision (AP) \cite{everingham2011pascal} as a metric to evaluate both object detection and instance segmentation performance. It summarizes the precision-recall curve at an intersection-over-union (IoU) threshold $\alpha$. For object detection, we use AP$^{box}$ to indicate the AP at a bounding box IoU between each predicted box and ground-truth box. For instance segmentation, we employ AP$^{mask}$ to represent the AP at mask IoU between each predicted segmentation mask and ground-truth mask. Following the previous works \cite{he2017mask,chen2019hybrid}, we report the averaged AP across the box or mask IoU thresholds from 0.5 to 0.95 with an interval of 0.05:
\begin{equation}
    \text{AP} = \frac{1}{10}\sum_{\alpha=0.5:0.05:0.95}\text{AP}_\alpha.
\end{equation}
We also present the AP$^{mask}$ at threshold $\alpha=0.5$ and $0.75$, following the work of \cite{he2017mask}. Besides, to show the quality of the segmentation, we report the averaged mask IoU AIoU$^{mask}_\alpha$ at $\alpha=0.5$ and $0.75$ :
\begin{equation}
    \text{AIoU}^{mask}_\alpha=\frac{1}{N}\sum_{i=1}^{N}\text{IoU}^{mask_i}_{\alpha},
\end{equation}
where $i$ indexes the $i$th instance object that has a mask IoU over threshold $\alpha$, and $N$ denotes the total number of such instances.

% Figure 4 %%%%%%%%%%%%%%%%%%%%%%%%%%%%%%%%%%%%%
\begin{figure*}[thb!]
\centering
\includegraphics[width=0.9\textwidth]{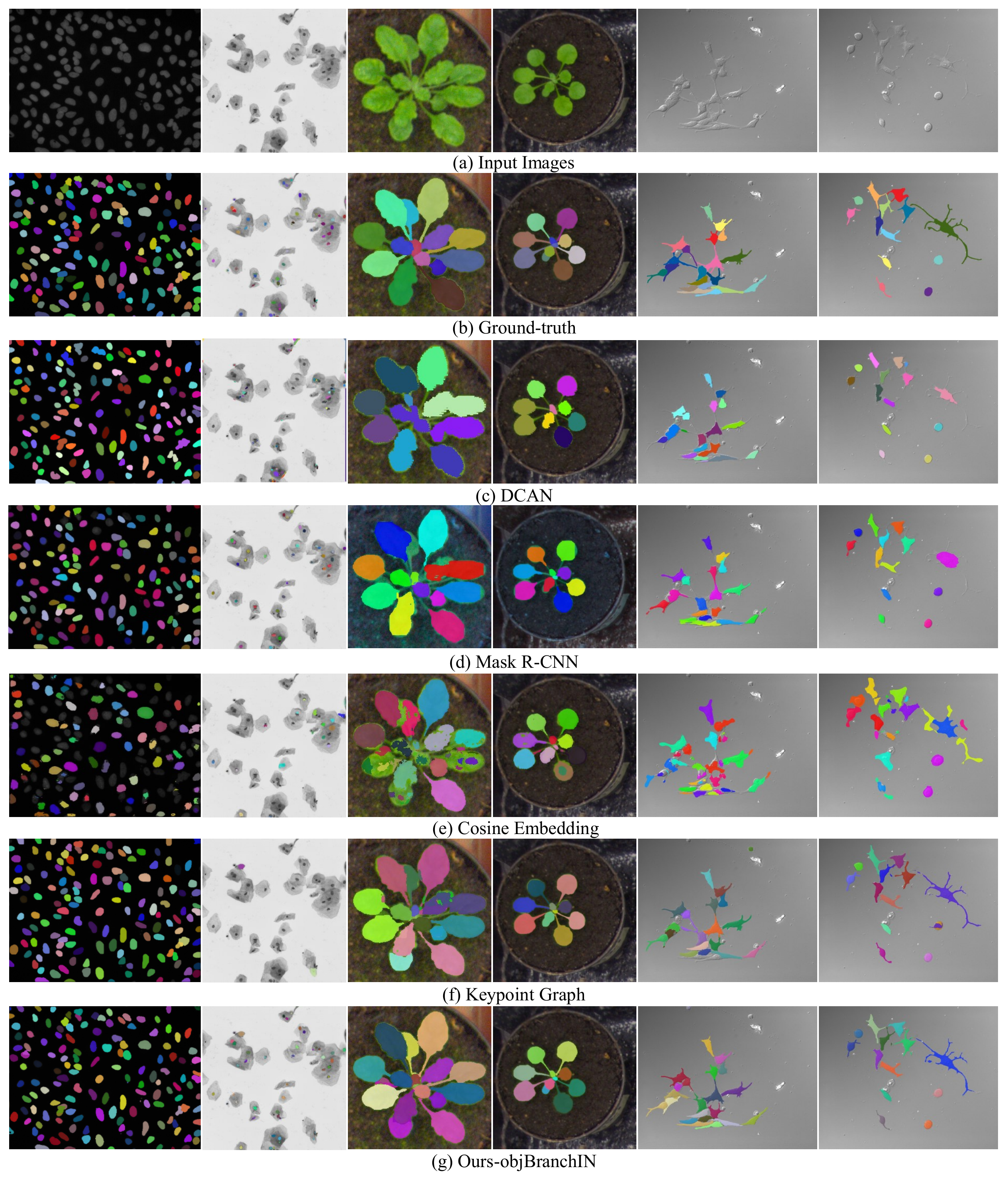}
\caption{Qualitative instance segmentation results. The ground-truth and predicted masks are overlayed on the input images.}
\label{fig3}
\end{figure*}
%%%%%%%%%%%%%%%%%%%%%%%%%%%%%%%%%%%%%%%%%%%%%%%%%%

\section{Experimental Results}
In this section, we compare our method with the following four works: box-free instance segmentation methods (DCAN \cite{chen2016dcan},  Cosine Embedding \cite{payer2018cosine}) and box-based instance segmentation methods (Mask R-CNN \cite{he2017mask}, Keypoint Graph \cite{YI2019MICCAI}). In addition, we perform several ablation studies to show the effectiveness of our object-guided segmentation branch in extracting the target details and suppressing the neighbor features.

\subsection{Comparison with State-of-the-arts}
The quantitative and qualitative instance segmentation results are shown in Table~\ref{table1} and Fig.~\ref{fig3}. In Table~\ref{table1}, $\text{AP}^{box}$ indicates the averaged detection performance over the bounding box IoU thresholds from 0.5 to 0.95 with an interval of 0.05. $\text{AP}^{mask}$ is the averaged instance segmentation performance over the same mask IoU thresholds. We explicitly exhibit $\text{AP}^{mask}$ at mask IoU threshold of 0.5 and 0.75. $\text{AIoU}^{mask}$ is used as an auxiliary metric when $\text{AP}^{mask}$ is too close for the compared methods. In Table~\ref{table1} it can be seen that although DCAN runs faster compared to the other baseline methods, its detection and instance segmentation accuracy are poor. The reason can be found from Fig.~\ref{fig3}c, where DCAN fails to separate the touching instances due to the unclear boundaries. Besides, by fusing the contours with the semantic maps, the instances would lose details such as the leaf stalks and cell protrusions. Cosine Embedding performs even worse compared to DCAN in Table~\ref{table1}. The reason would be that Cosine Embedding is likely to generate the mask fragments that belong to the same object (see Fig.~\ref{fig3}e). Compared to box-free instance segmentation methods, box-based instance segmentation methods excel in identifying individual objects. However, as can be seen from Fig.~\ref{fig3}d, Mask R-CNN is weak at detecting the objects that are very close to each other. This phenomenon may be caused by the sub-optimal training of the extremely imbalanced anchor-boxes. Besides, Mask R-CNN cannot correctly capture the object details such as the protrusion of cells and the leaf stalks due to the fixed-size of RoI patches. Keypoint Graph exhibits better detection ability compared to Mask R-CNN as it identifies the objects by grouping of keypoints. Besides, it is able to capture the object details due to the separate segmentation branch. In Table~\ref{table1}, we can see that the performance gap between Mask R-CNN and Keypoint Graph is small ($<$10 points) on DSB2018 and plant dataset. While on neural cell dataset with more slender and long structures, the gap gets bigger ($>$20 points). However, Keypoint Graph can't identify the small objects (see Fig.~\ref{fig3}f) due to the overlapping of the box keypoints. Also, it is unable to suppress the neighboring instances effectively. Compared to the methods mentioned above, the proposed object-guided instance segmentation method performs the best. From Fig.~\ref{fig3}g, we can see that the proposed method can identify small objects, retain object details, and suppress the neighbor noise features. Meanwhile, the proposed model is computationally efficient (see Table~\ref{table1}). These features demonstrate the superiority of the proposed method.

% Figure 3 %%%%%%%%%%%%%%%%%%%%%%%%%%%%%%%%%
\begin{figure*}[thb]
\centering
\includegraphics[width=0.9\textwidth]{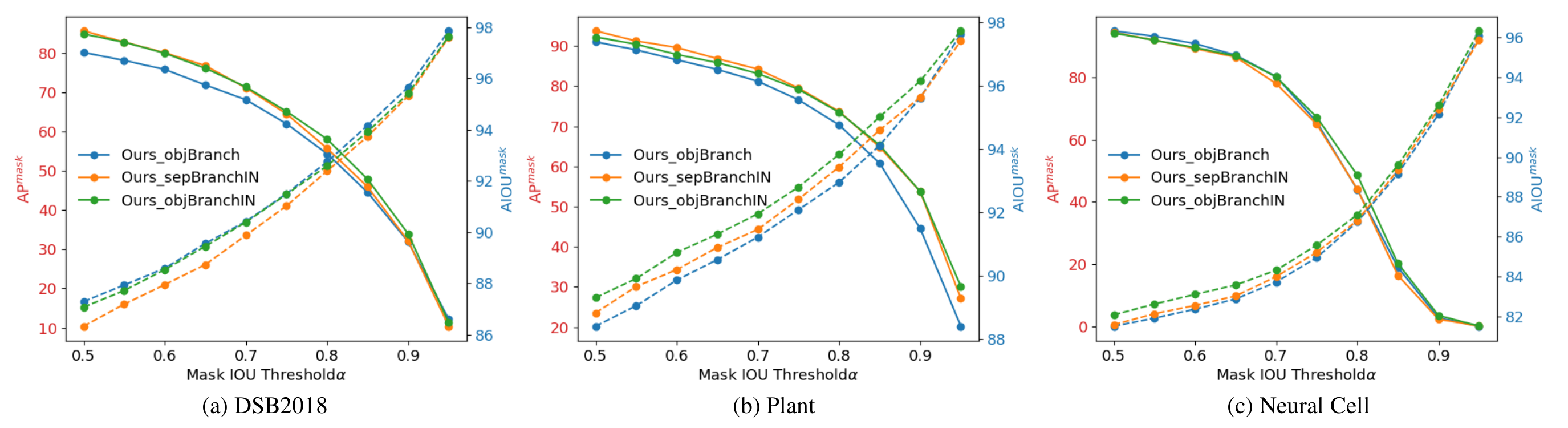}
\caption{Ablation studies of the segmentation branch for dataset DSB2018, plant phenotyping, and neural cells. The left axis is the AP$^{mask}$ along with the mask IoU thresholds ranging from 0.5 to 0.95 with an interval of 0.05. The right axis is the AIoU$^{mask}$. The solid lines indicate the AP$^{mask}$ and the dashed lines represent the AIoU$^{mask}$. The symbol ``obj" indicates the segmentation branch with the object feature, ``sep" denotes the separated segmentation branch that has no object features, and ``IN" refers to the instance normalization. AIoU$^{mask}$ is working as a secondary metric when AP$^{mask}$ is too close for the compared methods.}
\label{fig4}
\end{figure*}
%%%%%%%%%%%%%%%%%%%%%%%%%%%%%%%%%%%%%%%%%%%%

\subsection{Ablation Studies}
We perform ablation studies to show the effectiveness of our object-guided segmentation branch. The results are shown in Fig.~\ref{fig2} and Fig.~\ref{fig4}. The symbol ``sep" represents the separate segmentation without object features, ``IN" refers to the instance normalization, and ``obj" denotes the branch that has object features. As we can see from Fig.~\ref{fig2}, the separated segmentation branch (Ours\_sepBranchIN) can hardly differentiate the clustered instances due to a lack of object knowledge. In contrast, the segmentation branch with object features (Ours\_objBranch) can effectively separate the clustered objects. However, after the introduction of the object features, the predicted masks become incomplete. The reason would be that the object features are too coarse and they would perturb the distribution of the target instance. In addition, the neighboring features are not completely removed. This fact indicates that the model lacks the ability to filter out the neighbor information while retaining the target instance. With the instance normalization, the model (Ours\_objBranchIN) identifies the whole spatial information and is able to remove the neighbor statistics. Besides, it decreases the domination of the object features, making the optimal convergence possible. As a result, the predicted mask is intact and the neighbor noise features are suppressed. 

Fig.~\ref{fig4} compares the performances of segmentation branches at different mask IoU thresholds. For DSB2018 dataset (Fig.~\ref{fig4}a), Ours\_objBranch performs the worst in AP$^{mask}$. It would be caused by the fragmented neighboring instances and incomplete target segmentation (similar to Fig.~\ref{fig2}a). As a result, the number of unmatched pairs between the predicted instance masks and the ground-truth masks is increased. Ours\_sepBranchIN behaves better in AP$^{mask}$ as no object features are introduced to separate the neighbor instances. However, due to over-segmentation, the AIoU$^{mask}$ of Ours\_sepBranchIN is lower. Ours\_objBranchIN achieves the best results in both AP$^{mask}$ and AIoU$^{mask}$. For the plant phenotyping dataset (see Fig.~\ref{fig4}b), we can see that the Ours\_objBranch also performs the worst because of the same reason we explained above. The phenomenon suggests that the instance normalization is indispensable for the model to recover the intact instance details. Ours\_sepBranchIN and Ours\_objBranchIN performs close for AP$^{mask}$. At this time, we can check the segmentation quality from AIoU$^{mask}$. As can be seen from Fig.~\ref{fig4}b, the segmentation quality of Ours\_objBranchIN is consistently better than Ours\_sepBranchIN. For the neural cell dataset (see Fig.~\ref{fig4}c), Ours\_objBranchIN exhibits great superiority compared to Ours\_objBranch and Ours\_sepBranchIN. The reason would be that the neural cells generally attach, and they have long and slender protrusions, indicating  Ours\_objBranchIN is particularly good at dealing with clustered instances while keeping the instance details.

\subsection{Discussion for Multi-Modal Distribution}
%Q5: (1) How instance normalization generalizes to objects with multi-model appearance distribution? (2) How robust is the method for over-estimated object detections? A5: (1) 
Within an RoI patch, the multi-modal appearance distribution is generally mapped to multi-channels of the feature map. For some channels, the distribution of the target object could be weighted as the dominant one compared to the neighbors. In this paper, we first introduce the object feature to the segmentation branch. As the model can learn to adjust the instance normalization extent and the mapping weights for each channel, the model is able to suppress the neighbor distribution and highlight the target distribution under the ground-truth supervision. As a result, the proposed is excellent in separating the clustered objects and keeping the morphology details of the target object. Note that the detected bounding boxes in the instance segmentation are supposed to be tight. That's why object detection is crucial to box-based instance segmentation. Under this assumption, the proposed method is robust to recover the dominant distribution of the target object, which is verified by the experimental results and ablation studies. However, the proposed method would fail for an over-sized bounding box that contains several objects with identical sizes. But for an over-estimated bounding box contains only the target and background, the proposed method is still robust as the output probability feature contains classification information.

\section{Conclusion}
In this work, we propose a novel object-guided instance segmentation method that has three merits: (1) identifying small objects; (2) preserving the object details; (3) separating the clustered objects. The evaluation results of our method on the three biological datasets indicate that our model is particularly excellent in separating attached instances that need fine details. Compared to state-of-the-arts, the proposed method achieves favorable results in terms of both accuracy and speed on three biological datasets.
% Reference
%\bibliographystyle{aaai}
%\bibliography{AAAI-YiJ.2023}

\end{document}